\documentclass[10pt,twocolumn]{article}

\usepackage[utf8]{inputenc}
\usepackage[T1]{fontenc}
\usepackage{amsmath,amssymb}
\usepackage{graphicx}
\usepackage{booktabs}
\usepackage{multirow}
\usepackage{hyperref}
\usepackage{natbib}
\usepackage[margin=2cm]{geometry}
\usepackage{caption}
\usepackage{xcolor}
\usepackage{tikz}
\usetikzlibrary{arrows.meta, positioning, calc}

\title{Untrained CNNs Exceed Backpropagation in V1 Alignment at High Evaluation Resolution: A Systematic RSA Comparison of Four Learning Rules Against Human fMRI}

\author{Nils Leutenegger\\
Independent Researcher, Switzerland\\
\texttt{github.com/nilsleut}}

\date{}

\newcommand{\newid}{arXiv:2608.12408}

\input{numbers.tex}
\newcommand{\rs}[2]{\NV{rho.ps.cr.bn.#1.#2}}

\newcommand{\dps}[3]{\NV{d.ps.cr.bn.#1.#2.#3}}
\newcommand{\dpsci}[3]{[\NV{d.ps.cr.bn.#1.#2.#3.lo},\NV{d.ps.cr.bn.#1.#2.#3.hi}]}
\newcommand{\dmrci}[3]{[\NV{d.mr.cr.bn.#1.#2.#3.lo},\NV{d.mr.cr.bn.#1.#2.#3.hi}]}
\newcommand{\dpsciF}[3]{[\NV{d.ps.cr.bn.#1.#2.#3.lo.4},\NV{d.ps.cr.bn.#1.#2.#3.hi.4}]}
\newcommand{\dmrciF}[3]{[\NV{d.mr.cr.bn.#1.#2.#3.lo.4},\NV{d.mr.cr.bn.#1.#2.#3.hi.4}]}
\newcommand{\lb}[1]{\NV{lb.cr.#1}}
\newcommand{\lbci}[1]{[\NV{lb.cr.#1.lo},\NV{lb.cr.#1.hi}]}
\newcommand{\ml}[2]{\NV{ml.cr.bn.#1.#2}}
\newcommand{\mlci}[2]{[\NV{ml.cr.bn.#1.#2.lo},\NV{ml.cr.bn.#1.#2.hi}]}

\begin{document}
\maketitle

\begin{abstract}
\noindent\textbf{Revision note.} The central result of this paper, that the untrained network exceeds backpropagation in V1 alignment at 224 px, is not robust to the analysis choices examined in ``Analysis choices decide whether untrained networks beat trained ones at V1: a multiverse analysis of model--brain RSA'' arXiv:2611.15533. In the model set re-tested there, the V1 difference is $-$0.011 [$-$0.017, $-$0.005] with standardised model units, against +0.030 [0.018, 0.042] without, and a pre-specified ridge readout places backpropagation above the untrained network; under Euclidean distance the untrained advantage persists. This affects the title, the comparison of the untrained network with backpropagation at V1 and the conclusion on architecture. The backpropagation advantage at LOC holds with standardised units. The other analyses of this paper were not re-tested.

\medskip

A central question in computational neuroscience is whether the learning rule used to train a neural network determines how well its internal representations align with those of the human visual cortex. We present a systematic comparison of four learning rules---backpropagation (BP), feedback alignment (FA), predictive coding (PC), and spike-timing-dependent plasticity (STDP)---applied to identical convolutional architectures and evaluated against human fMRI data from the THINGS-fMRI dataset (720 stimuli, 3 subjects) using Representational Similarity Analysis (RSA). All models process stimuli at $224\times224$ resolution; results are averaged across 5 random seeds, and uncertainty is estimated by resampling stimuli. Crucially, we include an untrained random-weights baseline that isolates the contribution of architecture. At V1, the untrained baseline \emph{exceeds} backpropagation ($\rho = \rs{rnd}{V1}$ vs.\ $\rho = \rs{bp}{V1}$; $\Delta\rho = \dps{rnd}{bp}{V1}$, 95\% CI $\dpsci{rnd}{bp}{V1}$), and the same holds at V2. This advantage depends on the evaluation resolution: the models are trained at $32$\,px and evaluated at $224$\,px. In the 5-seed model set of the resolution study (same architecture and training; gap $\NV{res.gap.224.3}$ at $224$\,px), the gap vanishes when the models are evaluated at $32$\,px ($\Delta\rho = \NV{res.gap.32.3}$, 95\% CI $[\NV{res.gap.32.lo.3}, \NV{res.gap.32.hi.3}]$; \newid). At LOC, BP, PC and STDP exceed the random baseline, FA is borderline, and the trained rules do not differ from each other. At IT, no two conditions can be distinguished; the reliability of the IT data is low (leave-one-subject-out lower bound $\lb{IT}$). FA produces the lowest alignment at V1. Only the untrained network reaches the lower noise bound at V1; at V2 it and STDP are not distinguishable from the bound, and every condition lies below it at LOC and IT. These results indicate that, at the $224$\,px evaluation resolution used here, early visual alignment is driven more by the architecture than by the learning rule, and that learning-rule differences are resolvable only at intermediate areas.

\medskip
\noindent\textbf{Keywords:} representational similarity analysis, visual cortex, biologically plausible learning, random baseline, STDP, feedback alignment, predictive coding, fMRI, THINGS
\end{abstract}

\section{Introduction}

Deep neural networks (DNNs) trained with backpropagation have become the dominant computational models of the ventral visual stream \citep{yamins2016}. A well-established finding is that the hierarchy of features learned by DNNs mirrors the hierarchy of representations from V1 through IT cortex, as quantified by representational similarity analysis (RSA) and related metrics \citep{kriegeskorte2008, schrimpf2020}. This correspondence has been interpreted as evidence that the objectives and inductive biases of deep learning capture something fundamental about visual processing.

Yet backpropagation is neurobiologically implausible in at least three well-known respects: it requires symmetric forward and feedback weights (the weight transport problem), it relies on a global error signal propagated backward through the network, and the forward and backward passes must be temporally separated \citep{crick1989, lillicrap2020}. These observations have motivated work on biologically plausible alternatives, including feedback alignment \citep{lillicrap2016}, predictive coding \citep{rao1999, whittington2017}, and spike-timing-dependent plasticity \citep{bi1998}.

A key question is: does the choice of learning rule matter for brain alignment? If alignment arises from the architecture and objective rather than the weight-update rule, then alternative rules---or even no training at all---should yield comparable alignment. Prior work has compared individual learning rules to brain data \citep{schrimpf2020}, but to our knowledge no study has systematically compared all four rules together with an untrained random-weights baseline on the same architecture, stimuli, and RSA pipeline.

We address this gap with five conditions: BP, FA, PC, STDP, and a random-weights (untrained) baseline, all applied to identical CNNs trained on CIFAR-10, with layer-wise RDMs compared to human fMRI from the THINGS-fMRI dataset.

Our main contributions are: (1) a head-to-head RSA comparison of four learning rules plus a random baseline at $224\times224$ resolution, averaged across 5 seeds; (2) the finding that, at $224$\,px evaluation, the untrained CNN exceeds BP at V1 and V2 (V1: $\Delta\rho = \dps{rnd}{bp}{V1}$, 95\% CI $\dpsci{rnd}{bp}{V1}$), while at V1 the two do not differ at the $32$\,px training resolution; (3) the result that at IT, no two conditions can be distinguished at the low reliability of the IT data; (4) the finding that LOC is the only area where trained rules exceed the untrained baseline: BP, PC and STDP do so while the trained rules do not differ from each other; and (5) the finding that FA produces the lowest brain alignment at V1.

\section{Related Work}

\textbf{RSA as a model evaluation tool.} \citet{kriegeskorte2008} introduced RSA as a framework for comparing representations across measurement modalities. The core idea is to represent each system's response to a stimulus set as a pairwise dissimilarity matrix (RDM) and to assess model--brain alignment as the rank correlation between RDMs. RSA has since become a standard tool for comparing DNNs to neural data \citep{yamins2016, schrimpf2020}.

\textbf{DNNs and the visual hierarchy.} Task-optimized DNNs reproduce the representational hierarchy of the ventral visual stream, with early layers matching V1 and deeper layers matching IT \citep{yamins2016}. The Brain-Score benchmark \citep{schrimpf2020} systematizes this evaluation across architectures and datasets. Most high-scoring models are trained with standard BP on ImageNet. Importantly, \citet{schrimpf2020} noted that untrained networks retain some brain alignment, but did not systematically compare this against multiple learning rules.

\textbf{Biologically plausible learning rules.} Feedback alignment \citep{lillicrap2016} replaces the transpose weight matrix in the backward pass with fixed random matrices, removing the weight transport problem. \citet{lillicrap2016} showed that FA still allows learning in fully-connected networks, though its effectiveness in convolutional architectures is less established. Predictive coding \citep{rao1999} frames inference as hierarchical prediction-error minimization; \citet{whittington2017} showed that PC implements an approximation to BP under certain conditions, with learning driven entirely by local Hebbian updates. Spike-timing-dependent plasticity \citep{bi1998} updates synaptic weights based on the relative timing of pre- and postsynaptic spikes, and is one of the best-characterized forms of synaptic plasticity in mammalian cortex. Each of these rules has been studied individually in the context of neural representation learning, but direct comparisons on identical architectures and evaluation pipelines are rare.

\textbf{Random networks and early visual cortex.} \citet{saxe2011} demonstrated that random-weight networks can exhibit non-trivial visual structure. Concurrently with our work, \citet{truzzi2025} showed that random-weight CNNs predict early visual cortex responses when model complexity is sufficient. Our work extends these observations by systematically comparing four distinct learning rules against the random baseline across the visual hierarchy.

\textbf{Sparse coding and V1.} \citet{olshausen1996} showed that unsupervised learning of a sparse code for natural images produces receptive fields resembling V1 simple cells, establishing that correlation-based learning from image statistics suffices to produce V1-like representations without supervised objectives.

\section{Methods}

\subsection{Dataset}

We used the THINGS-fMRI dataset, part of THINGS-data \citep{hebart2023}: 3T fMRI responses (Siemens Magnetom Prisma) of 3 subjects to 8{,}740 unique images of 720 object concepts from the THINGS database \citep{hebart2019}, recorded in 12 main-experiment sessions of 10 runs each. We selected 720 stimuli (one image per concept, each presented once). fMRI responses were available for 3 subjects across 4 regions of interest (ROIs): V1, V2, LOC, and IT. For each subject and ROI, single-trial responses were z-scored per voxel across all trials, and a subject RDM was computed as the correlation distance between the response patterns of the 720 stimuli.

\subsection{Model Architecture}

All five conditions used an identical CNN architecture consisting of three convolutional layers (Conv1--Conv3, each followed by BatchNorm, ReLU, and $2\times2$ max-pooling) and two fully connected layers (FC1, FC2). The output layer uses softmax for 10-class classification on CIFAR-10. Training used 8,000 CIFAR-10 samples for 40 epochs at native $32\times32$ resolution. For RSA evaluation, THINGS stimuli were resized to $224\times224$ and passed through the same trained (or untrained) network; this is standard practice in model--brain comparisons where training and evaluation datasets differ \citep{schrimpf2020}. All results are from the five-seed re-run with the repaired evaluation mode (Revision note, item iii); the original model set is reported as a sensitivity column in the appendix.

The fixed layer-to-ROI mapping is Conv1$\to$V1, Conv1$\to$V2, Conv3$\to$LOC, FC1$\to$IT. V1 and V2 are both mapped to Conv1; Conv2 is not used in the fixed mapping (its values are given in the best-layer analysis, Table~\ref{tab:bestlayer}).

\begin{figure}[t]
\centering
\resizebox{\columnwidth}{!}{%
\begin{tikzpicture}[
    node distance=0.3cm,
    block/.style={rectangle, draw, rounded corners, minimum height=0.6cm, minimum width=0.9cm, font=\tiny, align=center, fill=blue!8},
    roi/.style={rectangle, draw, rounded corners, minimum height=0.4cm, minimum width=0.6cm, font=\tiny\bfseries, fill=green!15},
    arr/.style={-{Stealth[length=1.5mm]}, thick},
    darr/.style={-{Stealth[length=1.5mm]}, dashed, gray},
]
\node[font=\tiny\bfseries] at (3.0, 1.8) {5 Conditions: Random $\cdot$ BP $\cdot$ FA $\cdot$ PC $\cdot$ STDP};

\node[block] (input) at (0, 0.8) {Input\\$224{\times}224$};
\node[block, right=of input] (conv1) {Conv1\\BN+ReLU\\+Pool};
\node[block, right=of conv1] (conv2) {Conv2\\BN+ReLU\\+Pool};
\node[block, right=of conv2] (conv3) {Conv3\\BN+ReLU\\+Pool};
\node[block, right=of conv3] (fc1) {FC1\\512};
\node[block, right=of fc1, fill=orange!15] (fc2) {FC2\\Soft.};

\draw[arr] (input) -- (conv1);
\draw[arr] (conv1) -- (conv2);
\draw[arr] (conv2) -- (conv3);
\draw[arr] (conv3) -- (fc1);
\draw[arr] (fc1) -- (fc2);

\node[roi] (v1) at (1.5, -0.4) {V1};
\node[roi] (v2) at (2.7, -0.4) {V2};
\node[roi] (loc) at (3.9, -0.4) {LOC};
\node[roi] (it) at (5.1, -0.4) {IT};

\node[font=\tiny, gray] at (0.3, -0.4) {fMRI};
\node[font=\tiny, gray] at (3.3, -0.9) {early $\rightarrow$ late};

\draw[darr] (conv1.south) -- (v1.north);
\draw[darr] (conv1.south) -- (v2.north);
\draw[darr] (conv3.south) -- (loc.north);
\draw[darr] (fc1.south) -- (it.north);
\end{tikzpicture}%
}
\caption{Model architecture and layer-to-ROI mapping. Models are trained on CIFAR-10 ($32\times32$); for RSA, THINGS stimuli are resized to $224\times224$ at feature extraction time. Dashed arrows: Conv1$\to$V1, Conv1$\to$V2, Conv3$\to$LOC, FC1$\to$IT.}
\label{fig:architecture}
\end{figure}

\subsection{Learning Rules}

\textbf{Random Weights} serves as the architectural baseline. The network is initialized with PyTorch's default \texttt{kaiming\_uniform\_} and never trained, isolating the contribution of the convolutional architecture (filters, ReLU, pooling) to brain alignment. FA and STDP initialize with \texttt{kaiming\_normal\_}, whose standard deviation is larger by a factor of about $\NV{init.ratio}$. The untrained network therefore shares its initialization with BP and PC and is their matched control; for FA and STDP, comparisons with the untrained network also include the difference in initialization scale.

\textbf{Backpropagation (BP)} serves as the supervised baseline: $\Delta W_l = -\eta \frac{\partial \mathcal{L}}{\partial W_l}$.

\textbf{Feedback Alignment (FA)} \citep{lillicrap2016} replaces $\mathbf{W}_l^\top$ in the backward pass with a fixed random matrix $\mathbf{B}_l$:
\begin{equation}
\Delta W_l = \delta_l^{\text{FA}} \mathbf{h}_{l-1}^\top, \quad \delta_l^{\text{FA}} = f'(z_l) \odot (\mathbf{B}_{l+1} \delta_{l+1}^{\text{FA}}).
\end{equation}

\textbf{Predictive Coding (PC)} \citep{rao1999, whittington2017} minimizes a hierarchical prediction-error energy $F = \sum_l \|\varepsilon_l\|^2$ through iterative inference ($T_{\text{inf}} = 10$ steps, inference rate $\alpha = 0.02$):
\begin{equation}
\Delta W_l \propto \varepsilon_l \mathbf{x}_{l-1}^\top.
\end{equation}
Prediction weights $\mathbf{P}_l$ (implemented as transposed convolutions) generate top-down predictions; prediction errors $\varepsilon_l = r_l - \hat{r}_l$ drive both inference dynamics (updating representations $r_l$) and weight learning (weight rate $\eta_W = 10^{-4}$). As with STDP, only convolutional weights are updated via PC; the fully connected readout layers are trained with backpropagation.

\textbf{STDP} \citep{bi1998} converts activations to Poisson spike trains ($T = 10$ timesteps) and uses first-spike timing to update convolutional weights:
\begin{equation}
\Delta w = \begin{cases} A_+ e^{-\Delta t / \tau_+} & \Delta t > 0 \text{ (LTP)} \\ -A_- e^{\Delta t / \tau_-} & \Delta t < 0 \text{ (LTD)} \end{cases}
\end{equation}
with $\tau_+ = \tau_- = 20$\,ms, $A_+ = A_- = 0.003$, $\text{lr} = 5 \times 10^{-4}$. Only the convolutional layers (Conv1--Conv3) are updated via STDP; the fully connected layers (FC1, FC2) are trained with backpropagation to provide a supervised readout, analogous to a linear probe on STDP-learned features. The STDP test accuracy ($\NV{acc.stdp}\,\%$) therefore reflects STDP feature quality evaluated through a BP-trained classifier, not end-to-end STDP learning.

\subsection{RSA Pipeline}

\textbf{Feature extraction.} For each layer, activations were extracted on the 720 THINGS stimuli (resized to $224\times224$, standard ResNet input resolution). Convolutional activations were spatially averaged (global average pooling). All results are averaged across 5 random seeds per learning rule.

\textbf{RDM construction.} Model RDMs were computed as $1 - r$ (correlation distance).

\textbf{RSA score.} Model--brain alignment is Spearman's $\rho$ between the upper triangles of a model RDM and a subject RDM, restricted to the stimulus pairs defined below. The main convention ($\rho_{\text{subj}}$) correlates the model RDM with each subject's RDM and averages over the three subjects. The convention of earlier versions ($\rho_{\text{mean}}$) correlates the model RDM with the average of the three subject RDMs; it is reported alongside.

\textbf{Stimulus pairs (cross-run).} Each subject saw the 720 stimuli in ten runs, and sub-01 and sub-02 saw them in identical order. Responses within a run can share run-specific components (baseline drifts, run-wise noise), so for two subjects with the same order the dissimilarities of same-run pairs share a component that has nothing to do with the stimuli. This inflated the agreement between sub-01 and sub-02 (at LOC $\NV{agree.0102.LOC.all}$ on all pairs vs.\ $\NV{agree.0102.LOC.cr}$ on cross-run pairs, while sub-01 vs.\ sub-03 is $\NV{agree.0103.LOC.all}$ and $\NV{agree.0103.LOC.cr}$) and thereby the lower bound at LOC and IT (appendix, Table~\ref{tab:sens}). We therefore use only stimulus pairs that lie in different runs for all three subjects: $\NV{pairs.cr}$ of $\NV{pairs.total}$ pairs; $\NV{pairs.excl.pct}\,\%$ are excluded. Within a run, dissimilarity also depends on the distance between trials (Spearman correlation between dissimilarity and lag $\NV{lag.rho.min}$ to $\NV{lag.rho.max}$); at LOC and IT, neighbouring trials are on average anticorrelated (mean dissimilarity at lag 1 up to $\NV{lag.one.max}$). This would be consistent with collinearity between the overlapping hemodynamic responses of neighbouring trials in single-trial estimates, which is a further reason to exclude these pairs; we did not test this explanation. Excluding same-run pairs follows the same logic as run-wise cross-validation of dissimilarities, which keeps noise shared within a run out of the dissimilarity estimates \citep{walther2016}.

\textbf{Noise ceiling.} The lower bound follows \citet{nili2014}: each subject's RDM is correlated with the average RDM of the other two subjects, and the three correlations are averaged. It estimates how well a model that captures the structure shared across subjects would score, and is used here as a floor: a model below it leaves reliable shared structure unexplained. The upper bound (correlation of each subject with the average of all three, including itself) is not informative with three subjects: under a permutation null in which stimulus labels are shuffled independently per subject ($\NV{nperm}$ permutations), the upper bound stays large (V1: $\NV{ubnull.cr.V1.mean}$ against an observed $\NV{ub.cr.V1}$), because each subject contributes a third of the average it is compared with. It is therefore not shown in the figures. Under the same null the lower bound is $\NV{lbnull.cr.V1.mean} \pm \NV{lbnull.cr.V1.sd}$ (V1), so the observed lower bounds are far from chance at every ROI.

\textbf{Uncertainty.} 95\% confidence intervals come from a stimulus bootstrap: $\NV{nboot}$ resamples of the 720 stimuli with replacement (seed $\NV{bootseed}$), shared by all conditions, ROIs and quantities. Within each resample, pairs of a stimulus with itself are dropped and the cross-run pair set is recomputed. Differences between conditions, and between a condition and the lower bound, are computed on the same resamples, so their intervals account for the correlation between them. We call a difference significant when its percentile interval excludes 0. The three statements the conclusions rest on are treated as primary comparisons: the untrained network vs.\ BP at V1 and V2, BP vs.\ the untrained network at LOC, and FA vs.\ each other condition at V1 ($\NV{nprimary}$ of the 40 pairwise comparisons). They were designated in this revision, not before the original analysis. All other pairwise comparisons are exploratory. No correction for multiple testing is applied in the main text; all 40 intervals are reported, with the primary ones marked (appendix, Table~\ref{tab:pairwise}), and Benjamini--Hochberg and Holm corrections of the primary family are given in the appendix (Table~\ref{tab:fdr}).

\subsection{Partial RSA}

To control for low-level pixel similarity, a pixel RDM is computed as the pairwise correlation distance between the flattened RGB images (resized to $224$\,px on the shorter edge and centre-cropped to $224\times224$, as for the networks). Partial Spearman $\rho$ is estimated by residualizing the ranks of the model and the brain RDM vectors on the ranks of the pixel RDM vector and correlating the residuals. The pixel RDM was rebuilt for this version from the definition in the code of earlier versions; applied to the earlier analysis it reproduces the stored values to within $\NV{pval.dev}$ (they were stored at four decimals). Partial $\rho$ is computed per seed and subject on cross-run pairs and averaged as $\rho_{\text{subj}}$, with the same stimulus bootstrap.

\section{Results}

\subsection{Task Performance}

\begin{table}[h]
\centering
\caption{CIFAR-10 test accuracy ($\NV{acc.ntest}$ test images) of the repaired models after 40 epochs. Seed~0 only: checkpoints of the other four seeds were not stored, so no standard deviation can be given. Earlier versions reported training accuracy (appendix, Table~\ref{tab:accold}).}
\label{tab:accuracy}
\begin{tabular}{lc}
\toprule
Condition & Test accuracy (\%) \\
\midrule
Backprop & $75.0$ \\
STDP & $57.4$ \\
Predictive Coding & $56.7$ \\
Feedback Alignment & $38.8$ \\
Random Weights & $10.0$ \\
\bottomrule
\end{tabular}

\end{table}

\begin{figure*}[t]
\centering
\includegraphics[width=\textwidth]{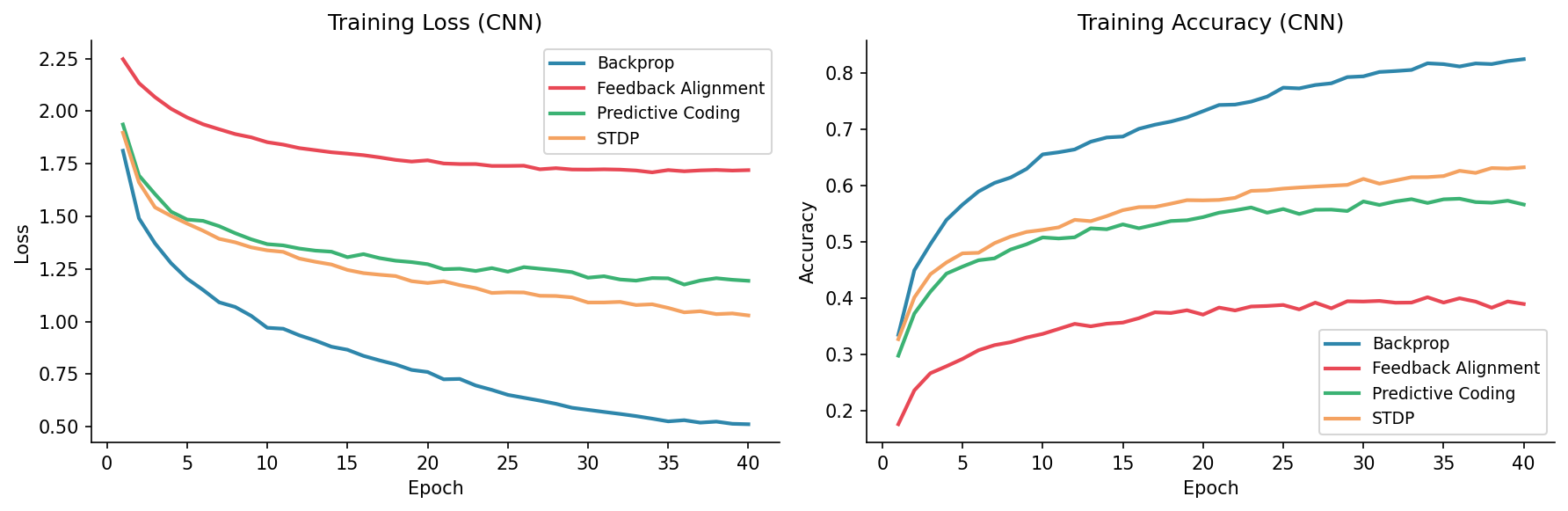}
\caption{Training dynamics of the original runs: loss (left) and accuracy (right) on the training batches over 40 epochs (8,000 CIFAR-10 samples) for all four learning rules. All runs are stable without divergence. Test accuracies of the repaired models are in Table~\ref{tab:accuracy}.}
\label{fig:training}
\end{figure*}

\subsection{Layer-wise Brain Alignment}

\begin{figure*}[t]
\centering
\includegraphics[width=\textwidth]{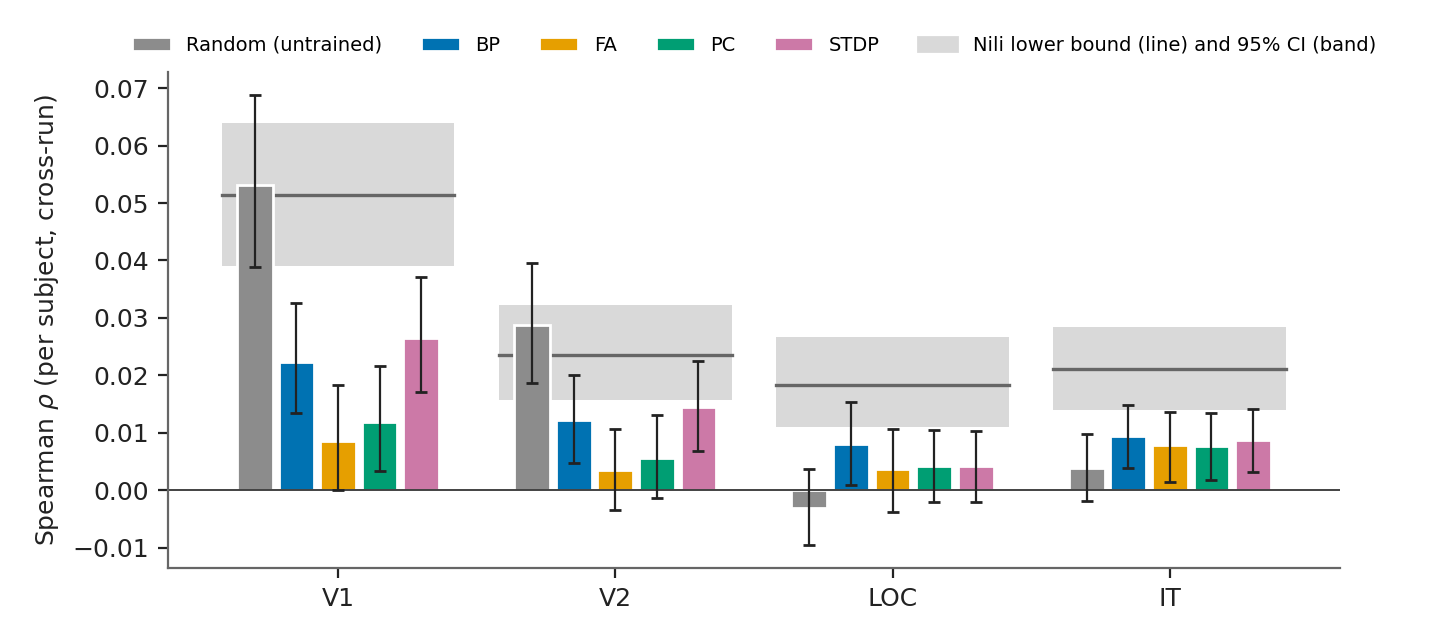}
\caption{Brain alignment across ROIs (main result). Spearman $\rho$ between model RDMs and each subject's RDM, averaged over the 3 subjects ($\rho_{\text{subj}}$), cross-run stimulus pairs, mean of 5 seeds. Error bars: stimulus-bootstrap 95\% CI. Grey line and band: \citet{nili2014} lower bound and its 95\% CI. The upper bound is not shown because it is not informative with three subjects (Methods).}
\label{fig:rsa_main}
\end{figure*}

\textbf{Interpreting absolute RSA scores.} Absolute Spearman $\rho$ values in this study are low (e.g., $\rho < 0.10$ at V1). They are not comparable with the scores of encoding-model benchmarks: RSA and encoding models measure fundamentally different quantities: RSA evaluates the geometry of the representational space (pairwise dissimilarity structure), whereas encoding models predict individual voxel responses, which is a more permissive criterion that can exploit stimulus-specific variance. RSA scores are further attenuated by the spatial resolution of fMRI relative to electrophysiology, and by our relatively small CNN architecture. The data themselves are of limited reliability: the lower bound is $\lb{V1}$ $\lbci{V1}$ at V1, $\lb{V2}$ at V2, $\lb{LOC}$ at LOC and $\lb{IT}$ at IT. Taken as a floor, it separates the conditions (model $-$ lower bound, with 95\% CI). At V1, only the untrained network reaches it ($\ml{rnd}{V1}$ $\mlci{rnd}{V1}$); every trained rule lies below it. At V2, the untrained network ($\ml{rnd}{V2}$ $\mlci{rnd}{V2}$) and STDP ($\ml{stdp}{V2}$ $\mlci{stdp}{V2}$) are not distinguishable from it, BP lies just below it (its interval ends less than $10^{-4}$ below zero), and FA and PC lie below it. At LOC and IT every condition lies below it. Reaching the lower bound is a weak criterion at V1: a single luminance value per image reaches $\rho_{\text{subj}} = \NV{lumv1.ps.cr}$ against the same V1 data, as much as the untrained network ($\rs{rnd}{V1}$; \newid).

Table~\ref{tab:rsa} provides the full numerical results. Several key patterns emerge.

\textbf{V1/V2---Untrained exceeds BP at 224\,px evaluation.} In early visual cortex, the untrained random-weights baseline \emph{exceeds} BP at V1 ($\rho = \rs{rnd}{V1}$ vs.\ $\rho = \rs{bp}{V1}$; $\Delta\rho = \dps{rnd}{bp}{V1}$ $\dpsci{rnd}{bp}{V1}$) and at V2 ($\rs{rnd}{V2}$ vs.\ $\rs{bp}{V2}$; $\Delta\rho = \dps{rnd}{bp}{V2}$ $\dpsci{rnd}{bp}{V2}$). The order at V1 is Random $>$ STDP $>$ BP $>$ PC $>$ FA, and all ten pairwise differences at V1 are significant in the per-subject convention; in the mean-RDM convention BP and STDP do not differ ($\dmrciF{bp}{stdp}{V1}$), and at V2 they do not differ in either convention. At this evaluation resolution, the convolutional architecture---local connectivity, ReLU, and pooling---is thus the primary driver of early visual alignment. When the models are evaluated at their $32$\,px training resolution, BP and the untrained network do not differ at V1 ($\rho_{\text{subj}} = \NV{res.bp.32}$ vs.\ $\NV{res.rnd.32}$; $\Delta\rho = \NV{res.gap.32}$, 95\% CI $[\NV{res.gap.32.lo},\NV{res.gap.32.hi}]$; cross-run pairs, model set of the resolution study, same architecture and training, 5 seeds); the untrained network's advantage at $224$\,px arises from the difference between training and evaluation resolution (\newid).

\textbf{LOC---Learning effect without differences among rules.} At LOC, BP ($\rs{bp}{LOC}$), PC ($\rs{pc}{LOC}$) and STDP ($\rs{stdp}{LOC}$) exceed the random baseline ($\rs{rnd}{LOC}$; BP $-$ Random $\dps{bp}{rnd}{LOC}$ $\dpsci{bp}{rnd}{LOC}$, PC $-$ Random $\dps{pc}{rnd}{LOC}$ $\dpsci{pc}{rnd}{LOC}$, STDP $-$ Random $\dps{stdp}{rnd}{LOC}$ $\dpsci{stdp}{rnd}{LOC}$). FA is borderline: its difference from Random is $\NV{d.ps.cr.bn.fa.rnd.LOC.4}$ $\dpsciF{fa}{rnd}{LOC}$ in the per-subject convention and excludes 0 in the mean-RDM convention $\dmrci{fa}{rnd}{LOC}$. No difference between two trained rules is significant (e.g.\ BP $-$ FA $\dps{bp}{fa}{LOC}$ $\dpsci{bp}{fa}{LOC}$).

\textbf{IT---Not distinguishable at low reliability.} At IT, the five conditions range from $\NV{range.ps.IT.min}$ to $\NV{range.ps.IT.max}$, and none of the ten pairwise differences is significant, including Random vs.\ BP ($\dps{rnd}{bp}{IT}$ $\dpsci{rnd}{bp}{IT}$). Reliability at IT is low, second lowest after LOC (lower bound $\lb{IT}$ $\lbci{IT}$), and every condition lies below this bound. The conditions are therefore not distinguishable at the reliability available at IT; this does not show that they are equivalent.

\textbf{FA lowest at V1.} FA produces the lowest brain alignment at V1 ($\rs{fa}{V1}$), and its differences from every other condition at V1 are significant in both conventions, suggesting that random feedback filters disrupt the representational structure that the convolutional architecture provides for free. This holds despite FA reaching $\NV{acc.fa}\,\%$ test accuracy---substantially above chance---indicating that FA learns task-relevant features that do not correspond to human visual representations. At V2, the ordering among FA, BP and PC depends on the layer assignment (Conv1 vs.\ Conv2; Appendix Table~\ref{tab:v2conv2}), whereas Random $>$ BP holds under both.

\begin{table*}[t]
\centering
\caption{RSA scores $\rho_{\text{subj}}$ (per-subject convention; mean across 5 seeds) with stimulus-bootstrap 95\% CI, cross-run stimulus pairs, repaired model set, $224\times224$ input. The mean-RDM convention is in the appendix (Table~\ref{tab:rsa_mr}). Lower bound: \citet{nili2014} leave-one-subject-out. Bold: highest $\rho_{\text{subj}}$ per ROI (V1, V2, LOC). $^{\dagger}$: difference from Random Weights significant (CI excludes 0). $^{\ddagger}$: at IT no pairwise difference is significant; no condition is marked.}
\label{tab:rsa}
\small
\resizebox{\textwidth}{!}{%
\begin{tabular}{llcccccc}
\toprule
ROI & Layer & lower bound & Random & BP & FA & PC & STDP \\
\midrule
V1 & Conv1 & $0.051\;[.039,.064]$ & $\mathbf{0.053}\;[.039,.069]$ & $0.022\;[.013,.033]^{\dagger}$ & $0.009\;[.000,.018]^{\dagger}$ & $0.012\;[.003,.022]^{\dagger}$ & $0.027\;[.017,.037]^{\dagger}$ \\
V2 & Conv1 & $0.024\;[.016,.032]$ & $\mathbf{0.029}\;[.019,.040]$ & $0.012\;[.005,.020]^{\dagger}$ & $0.004\;[-.003,.011]^{\dagger}$ & $0.006\;[-.001,.013]^{\dagger}$ & $0.015\;[.007,.023]^{\dagger}$ \\
LOC & Conv3 & $0.018\;[.011,.027]$ & $-0.003\;[-.010,.004]$ & $\mathbf{0.008}\;[.001,.015]^{\dagger}$ & $0.004\;[-.004,.011]$ & $0.004\;[-.002,.011]^{\dagger}$ & $0.004\;[-.002,.010]^{\dagger}$ \\
IT & FC1 & $0.021\;[.014,.028]$ & $0.004\;[-.002,.010]$ & $0.009\;[.004,.015]$ & $0.008\;[.001,.014]$ & $0.008\;[.002,.013]$ & $0.009\;[.003,.014]$$^{\ddagger}$ \\
\bottomrule
\end{tabular}
}%
\end{table*}

\subsection{Hierarchical Gradient}

\begin{figure}[t]
\centering
\includegraphics[width=\columnwidth]{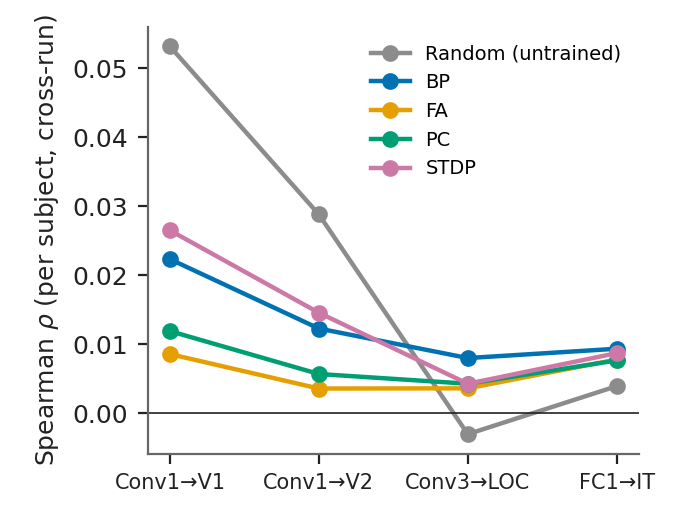}
\caption{Brain alignment along the fixed layer-to-ROI mapping ($\rho_{\text{subj}}$, cross-run). All conditions are lower at LOC than at V1. At IT the conditions lie between $\NV{range.ps.IT.min}$ and $\NV{range.ps.IT.max}$ and are not distinguishable. The untrained network is the highest at V1/V2 and the lowest at LOC and IT.}
\label{fig:hierarchy}
\end{figure}

All five conditions are lower at LOC than at V1 (Figure~\ref{fig:hierarchy}). The random baseline shows the most distinctive pattern: highest at V1/V2 (exceeding all trained rules), dropping below zero at LOC ($\rs{rnd}{LOC}$, the lowest), and at IT ($\rs{rnd}{IT}$) the lowest again but not significantly different from any trained rule. Convolutional inductive biases thus drive early alignment, while at IT the differences between conditions are below what the data can resolve.

\subsection{Pairwise Comparisons}

All 40 pairwise differences (10 pairs $\times$ 4 ROIs) with their bootstrap intervals are listed in Table~\ref{tab:pairwise}. At V1, all ten are significant; at V2, all except BP vs.\ STDP. At LOC, the differences between the random baseline and BP, PC and STDP are significant, Random vs.\ FA is borderline, and none of the six differences among trained rules is significant. At IT, none of the ten is significant.

\subsection{Subject-Level Consistency}

\begin{figure*}[t]
\centering
\includegraphics[width=\textwidth]{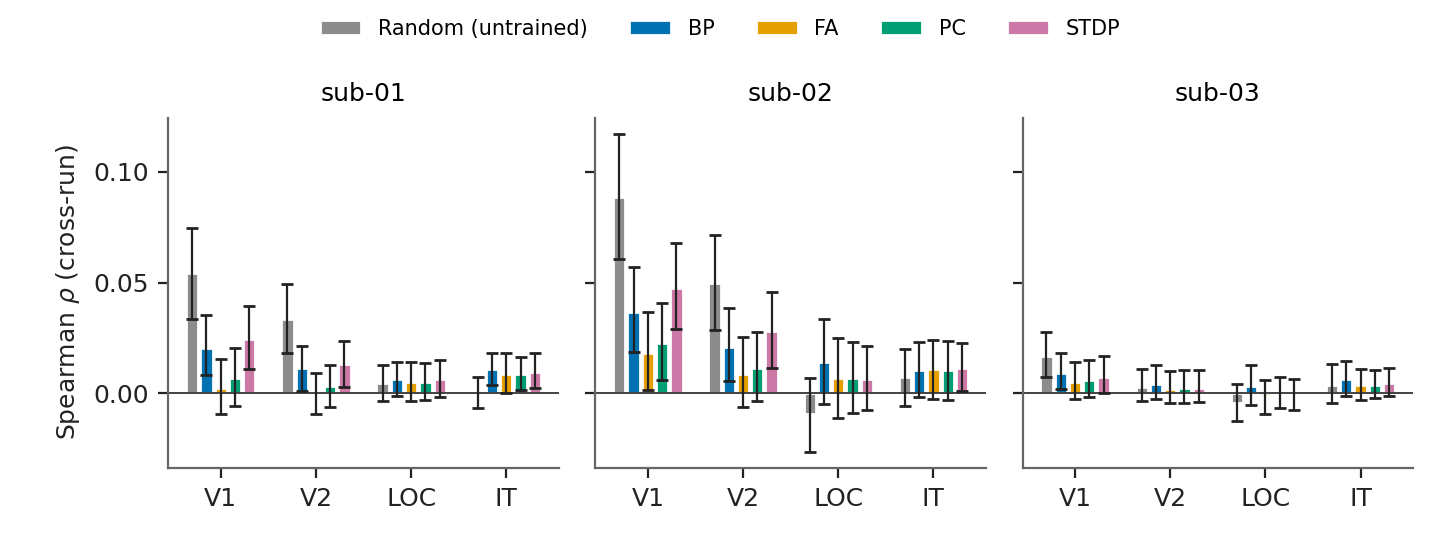}
\caption{Single-subject RSA scores per learning rule and ROI (cross-run pairs, mean of 5 seeds), with stimulus-bootstrap 95\% CIs. Sub-03 shows weaker alignment than sub-01 and sub-02 across conditions and ROIs. In sub-01 and sub-02 the V1 order of the point estimates is that of the group; at LOC, no difference between two trained rules is significant in any single subject.}
\label{fig:subjects}
\end{figure*}

We assessed whether the results hold within individual subjects (Figure~\ref{fig:subjects}; single-subject $\rho$ with stimulus-bootstrap CIs on the same resamples). Sub-03 shows weaker alignment across all conditions and ROIs, suggesting lower fMRI signal quality. At V1, sub-01 and sub-02 each show the group order Random $>$ STDP $>$ BP $>$ PC $>$ FA in their point estimates, and in each of them the untrained network exceeds every trained rule (e.g.\ Random $-$ BP: sub-01 $\NV{subd.s1.rnd.bp.V1}$ $[\NV{subd.s1.rnd.bp.V1.lo},\NV{subd.s1.rnd.bp.V1.hi}]$, sub-02 $\NV{subd.s2.rnd.bp.V1}$ $[\NV{subd.s2.rnd.bp.V1.lo},\NV{subd.s2.rnd.bp.V1.hi}]$). FA is below every other condition in sub-01; in sub-02 its difference from PC is borderline. At LOC, single subjects do not resolve the differences: BP has the highest point estimate in sub-02 and sub-03, and in sub-01 STDP ($\NV{sub.s1.stdp.LOC}$) is marginally above BP ($\NV{sub.s1.bp.LOC}$); no difference between two trained rules is significant in any subject, and the difference between the untrained network and BP is significant only in sub-02 (Random $-$ BP $\NV{subd.s2.rnd.bp.LOC}$ $[\NV{subd.s2.rnd.bp.LOC.lo},\NV{subd.s2.rnd.bp.LOC.hi}]$). The LOC result of the main analysis therefore rests on the combination of subjects, not on a consistent per-subject ranking.

\subsection{Seed Variability}

\begin{figure*}[t]
\centering
\includegraphics[width=\textwidth]{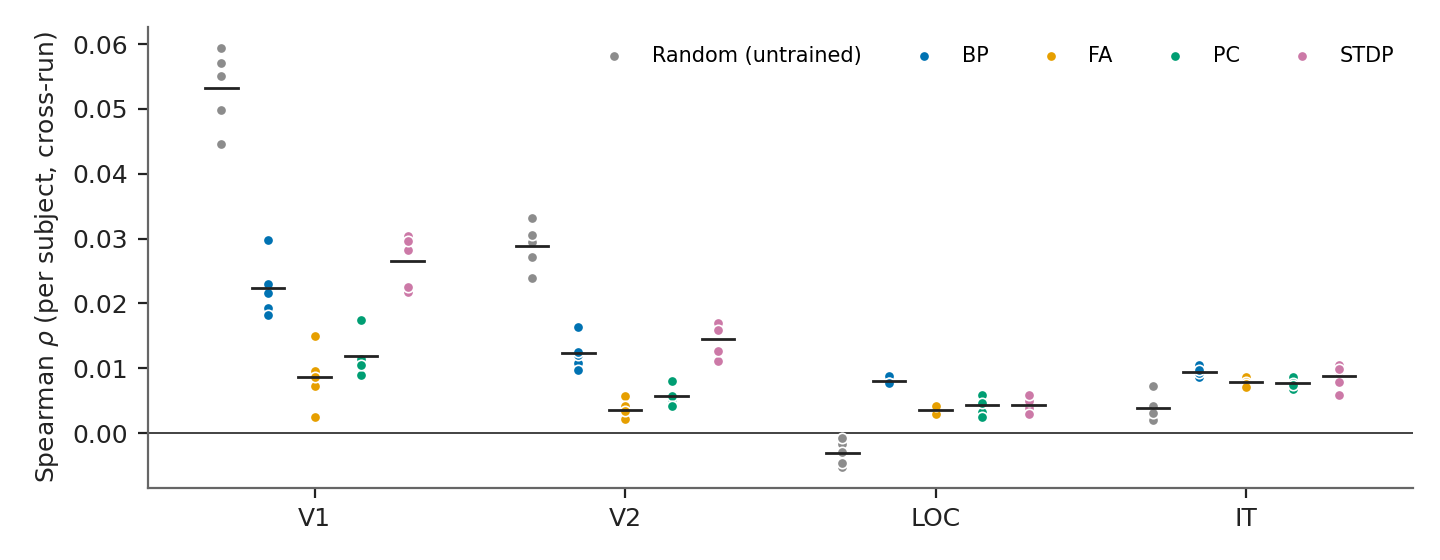}
\caption{RSA scores of the 5 seeds per condition (dots) and their mean (line), $\rho_{\text{subj}}$, cross-run pairs. At V1/V2, between-rule differences are large relative to the seed spread. At IT, the seed spread is small, and the lack of significant differences reflects the stimulus-sampling uncertainty.}
\label{fig:seeds}
\end{figure*}

To assess robustness to random initialization, we trained each learning rule with 5 different random seeds (Figure~\ref{fig:seeds}). At V1/V2, the seed standard deviation ($\NV{seedsd.v12.min}$--$\NV{seedsd.v12.max}$) is small relative to the differences between conditions (e.g.\ $\Delta\rho = \dps{rnd}{bp}{V1}$ between Random and BP at V1), so the order of conditions is stable across initializations. At IT, the seed standard deviations are also small ($\NV{seedsd.it.min}$--$\NV{seedsd.it.max}$), and most seed ranges overlap; BP's lowest seed ($\NV{seed5.bp.IT.min}$) lies just above FA's highest ($\NV{seed5.fa.IT.max}$) and above the untrained network's highest ($\NV{seed5.rnd.IT.max}$). The seed spread does not include the uncertainty from the sampling of stimuli, which is much larger (Random $-$ BP at IT: $\dpsci{rnd}{bp}{IT}$); it is this uncertainty that leaves the IT comparisons unresolved.

\subsection{Best-Layer-per-ROI Analysis}

We evaluated whether the fixed layer-to-ROI mapping is optimal by computing RSA for all layer$\times$ROI combinations (Table~\ref{tab:bestlayer}; point estimates, $\rho_{\text{subj}}$, cross-run). For BP, the fixed mapping matches the best layer at $\NV{bl.bp.nmatch}$ of four ROIs; at LOC, FC1 is higher than Conv3. For Random, FA and PC, FC1 produces the highest V1 alignment (Random: $\NV{bl.rnd.V1}$ vs.\ $\NV{blfix.rnd.V1}$ at Conv1)---likely reflecting that FC1 compresses all information into a 512-dimensional space that happens to correlate with the fMRI signal. This finding suggests that best-layer analysis can be misleading when layer dimensionality varies, and supports the use of fixed anatomically-motivated mappings.

\begin{table}[h]
\centering
\caption{Best layer per ROI and its $\rho_{\text{subj}}$ (cross-run, mean of 5 seeds; point estimates). In parentheses: the fixed layer's value where it differs.}
\label{tab:bestlayer}
\small
\resizebox{\columnwidth}{!}{%
\begin{tabular}{lcccc}
\toprule
Condition & V1 & V2 & LOC & IT \\
\midrule
Random & FC1 $0.102$ ($0.053$) & FC1 $0.061$ ($0.029$) & Conv1 $-0.002$ ($-0.003$) & FC1 $0.004$ \\
BP & Conv1 $0.022$ & Conv1 $0.012$ & FC1 $0.012$ ($0.008$) & FC1 $0.009$ \\
FA & FC1 $0.020$ ($0.009$) & FC1 $0.014$ ($0.004$) & FC1 $0.005$ ($0.004$) & FC1 $0.008$ \\
PC & FC1 $0.015$ ($0.012$) & FC1 $0.013$ ($0.006$) & Conv2 $0.005$ ($0.004$) & FC1 $0.008$ \\
STDP & Conv2 $0.027$ ($0.027$) & FC1 $0.018$ ($0.015$) & FC1 $0.007$ ($0.004$) & FC1 $0.009$ \\
\bottomrule
\end{tabular}
}
\end{table}

\subsection{Partial RSA}

Removing the variance explained by pixel-level similarity lowers every condition's V1 score (\(\Delta\) between $\NV{pr.d.V1.min}$ and $\NV{pr.d.V1.max}$), and each of these decreases is significant (Table~\ref{tab:partial}). The order at V1 is unchanged (Random $>$ STDP $>$ BP $>$ PC $>$ FA), and the untrained network still exceeds BP at V1 ($\NV{prd.rnd.bp.V1}$ $[\NV{prd.rnd.bp.V1.lo},\NV{prd.rnd.bp.V1.hi}]$) and V2 ($\NV{prd.rnd.bp.V2}$ $[\NV{prd.rnd.bp.V2.lo},\NV{prd.rnd.bp.V2.hi}]$). Across all 40 pairwise comparisons, the same differences are significant with and without pixel control, so none of the results reported above is carried by pixel-level similarity.

\begin{table*}[t]
\centering
\caption{Standard ($\rho_{\text{subj}}$) and partial RSA ($\rho_{\text{partial}}$, pixel similarity controlled) at V1 and V2; $\Delta = \rho_{\text{partial}} - \rho_{\text{subj}}$. Cross-run pairs, repaired model set, per-subject convention, mean of 5 seeds, stimulus-bootstrap 95\% CI.}
\label{tab:partial}
\small
\resizebox{\textwidth}{!}{\begin{tabular}{lcccccc}
\toprule
 & \multicolumn{3}{c}{V1} & \multicolumn{3}{c}{V2} \\
\cmidrule(lr){2-4} \cmidrule(lr){5-7}
Condition & $\rho_{\text{subj}}$ & $\rho_{\text{partial}}$ [95\% CI] & $\Delta$ [95\% CI] & $\rho_{\text{subj}}$ & $\rho_{\text{partial}}$ [95\% CI] & $\Delta$ [95\% CI] \\
\midrule
Random & $0.053$ & $0.051\;[0.037,0.066]$ & $-0.0025\;[-0.0042,-0.0011]$ & $0.029$ & $0.028\;[0.018,0.038]$ & $-0.0011\;[-0.0024,-0.0000]$ \\
STDP & $0.027$ & $0.022\;[0.013,0.032]$ & $-0.0045\;[-0.0069,-0.0024]$ & $0.015$ & $0.013\;[0.005,0.020]$ & $-0.0020\;[-0.0041,-0.0002]$ \\
BP & $0.022$ & $0.017\;[0.009,0.027]$ & $-0.0052\;[-0.0080,-0.0027]$ & $0.012$ & $0.010\;[0.003,0.017]$ & $-0.0022\;[-0.0045,-0.0002]$ \\
PC & $0.012$ & $0.007\;[-0.001,0.017]$ & $-0.0046\;[-0.0071,-0.0026]$ & $0.006$ & $0.004\;[-0.003,0.011]$ & $-0.0021\;[-0.0041,-0.0005]$ \\
FA & $0.009$ & $0.004\;[-0.004,0.013]$ & $-0.0045\;[-0.0069,-0.0024]$ & $0.004$ & $0.002\;[-0.005,0.008]$ & $-0.0020\;[-0.0040,-0.0003]$ \\
\bottomrule
\end{tabular}
}
\end{table*}

\subsection{Filter Analysis}

\begin{figure}[t]
\centering
\includegraphics[width=\columnwidth]{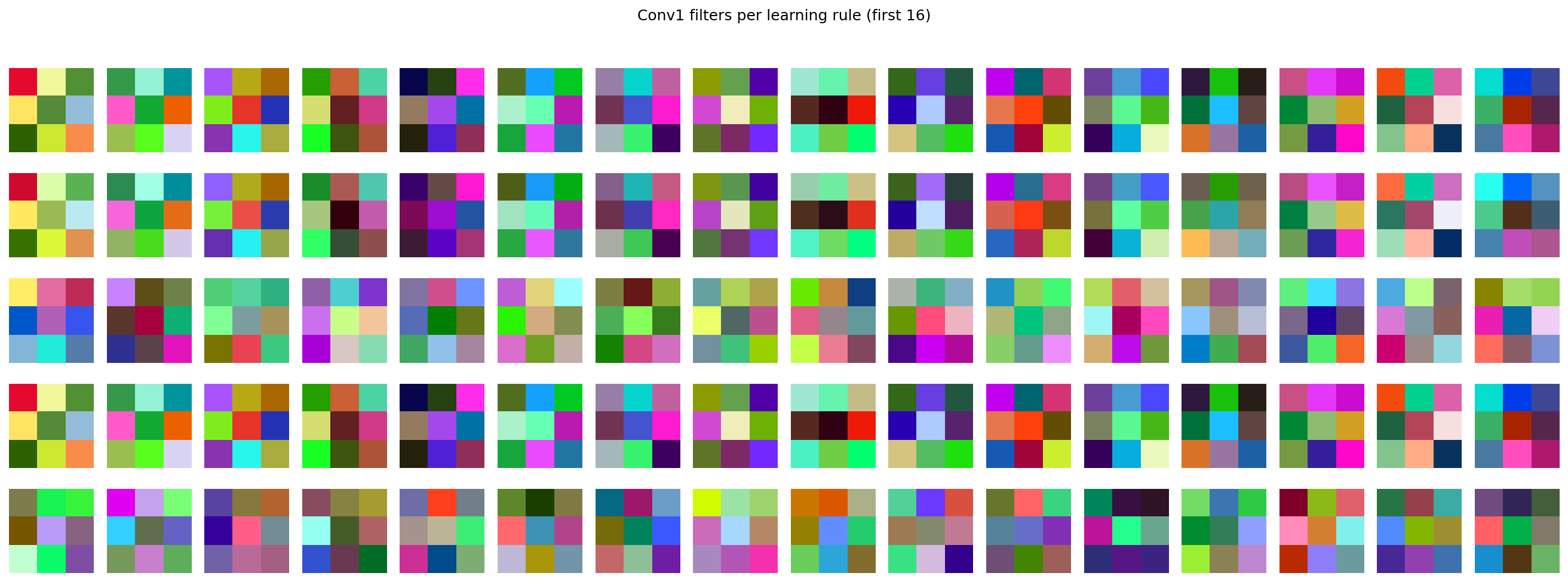}
\vspace{0.3cm}
\includegraphics[width=\columnwidth]{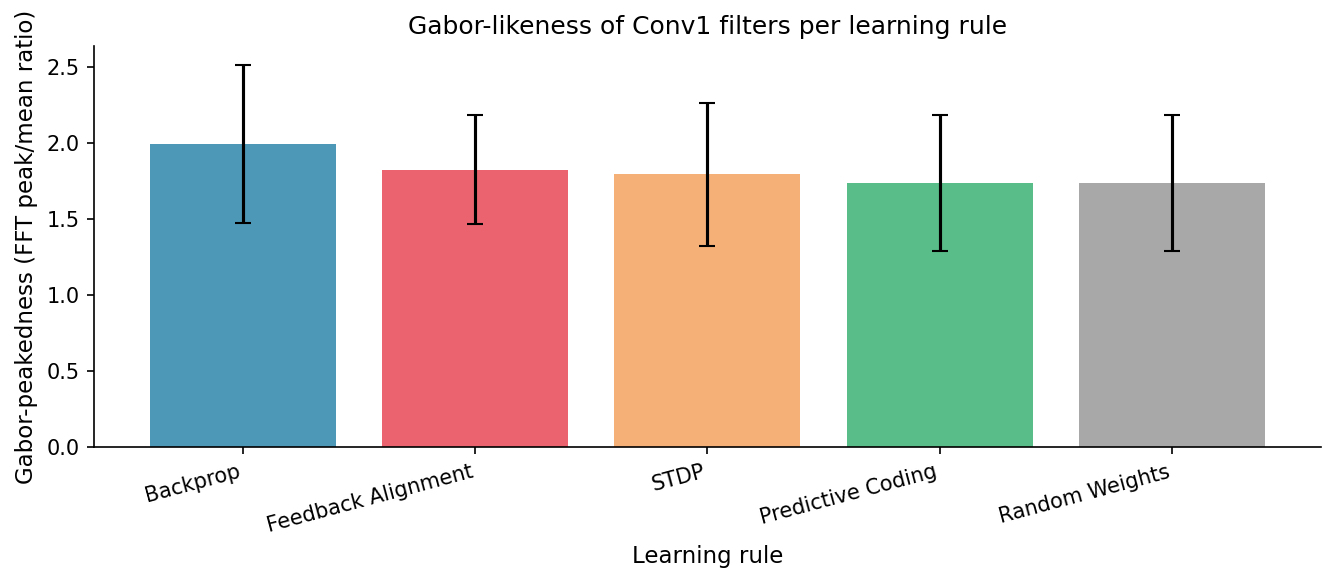}
\caption{Conv1 filter analysis for all five conditions (original model set; weight-based, so not affected by the evaluation defect). (a) First 16 Conv1 filters per learning rule. All conditions show color-opponent patch structure rather than oriented Gabor patterns, consistent with CIFAR-10 training at $32\times32$. (b) Gabor-peakedness (FFT spectral peak-to-mean ratio): BP $\NV{gabor.bp.mean} \pm \NV{gabor.bp.std}$, FA $\NV{gabor.fa.mean} \pm \NV{gabor.fa.std}$, STDP $\NV{gabor.stdp.mean} \pm \NV{gabor.stdp.std}$, PC $\NV{gabor.pc.mean} \pm \NV{gabor.pc.std}$, Random $\NV{gabor.rnd.mean} \pm \NV{gabor.rnd.std}$. Differences are small relative to within-rule variability.}
\label{fig:filters}
\end{figure}

Conv1 filters from all five conditions exhibit color-opponent patch structure rather than oriented Gabor-like patterns, consistent with CIFAR-10 training. The Gabor-peakedness score (FFT spectral peak-to-mean ratio) is highest for BP ($\NV{gabor.bp.mean} \pm \NV{gabor.bp.std}$), followed by FA ($\NV{gabor.fa.mean} \pm \NV{gabor.fa.std}$), STDP ($\NV{gabor.stdp.mean} \pm \NV{gabor.stdp.std}$), PC ($\NV{gabor.pc.mean} \pm \NV{gabor.pc.std}$), and Random ($\NV{gabor.rnd.mean} \pm \NV{gabor.rnd.std}$). The differences are small relative to within-rule variability and standard deviations overlap substantially across all conditions, indicating that no learning rule produces markedly more Gabor-like filters than random initialization at this scale.

\section{Discussion}

\textbf{Architecture as the primary driver of early alignment.} Our most striking finding is that an untrained CNN \emph{exceeds} backpropagation at V1 ($\rho = \rs{rnd}{V1}$ vs.\ $\rs{bp}{V1}$, $\Delta\rho = \dps{rnd}{bp}{V1}$ $\dpsci{rnd}{bp}{V1}$). This goes beyond the observation of \citet{saxe2011} that random networks have non-trivial visual structure: in our setup, at $224$\,px evaluation, the convolutional inductive biases (local connectivity, ReLU, pooling) are not merely comparable to BP but produce higher early visual alignment. The advantage is not evidence that training degrades early-layer structure: when the models are evaluated at their $32$\,px training resolution, BP and the untrained network do not differ at V1 ($\rho_{\text{subj}} = \NV{res.bp.32}$ vs.\ $\NV{res.rnd.32}$; $\Delta\rho = \NV{res.gap.32}$, 95\% CI $[\NV{res.gap.32.lo},\NV{res.gap.32.hi}]$, model set of the resolution study), and the untrained network's lead at $224$\,px arises from the difference between training and evaluation resolution (\newid). Whether the same holds at larger scales (e.g., ResNet-50 on ImageNet) remains to be tested.

\textbf{On the untrained baseline.} The untrained condition's batch-normalisation layers hold their initialised statistics, so in evaluation mode they act as the identity, whereas every trained condition carries statistics accumulated during training. A $2\times2$ calibration control varying the calibration set and the calibration resolution, with the convolutional weights held bit-identical, indicates that this does not produce the result: with statistics calibrated at the evaluation resolution the untrained network still exceeds backpropagation at V1 by $\NV{cal.B}$ (CIFAR-10 statistics) or $\NV{cal.D}$ (statistics from a disjoint pool of THINGS images), $\NV{cal.B.n}$ and $\NV{cal.D.n}$ seeds (original analysis: all pairs, mean-RDM convention). The result is, however, strongly dependent on the evaluation resolution, which this paper holds fixed at $224$\,px; see \newid.

\textbf{Few reliable differences at higher areas.} At LOC, BP, PC and STDP exceed the random baseline, FA is borderline, and the differences among trained rules are small and not significant. At IT, no two conditions can be distinguished. This may reflect a genuine property of abstract categorical representations, but is also consistent with a capacity limitation: our small CNN trained on 8,000 CIFAR-10 samples may simply lack the representational power to differentiate at higher layers. Prior work with larger models suggests that BP-trained networks have a systematic IT advantage \citep{schrimpf2020}; whether this advantage reflects the learning rule or the additional model capacity remains an open question. The low reliability of the IT data (lower bound $\lb{IT}$, with every condition below it) also leaves little room for differences to be resolved with three subjects; true differences may exist but are not resolvable here.

\textbf{FA lowest at V1.} FA produces the lowest alignment at V1, significantly below all other conditions, despite reaching $\NV{acc.fa}\,\%$ test accuracy. This is consistent with random feedback filters disrupting the representational structure that convolutional architecture provides, though we cannot rule out that this reflects an optimization artifact specific to FA in convolutional networks rather than a fundamental incompatibility. After pixel control, FA's V2 score is $\NV{pr.partial.fa.V2}$ $[\NV{pr.partial.fa.V2.lo},\NV{pr.partial.fa.V2.hi}]$; it was already not distinguishable from zero without pixel control ($\NV{pr.std.fa.V2}$ $[\NV{pr.std.fa.V2.lo},\NV{pr.std.fa.V2.hi}]$). This result may be specific to CNNs, where spatial structure in feature maps interacts poorly with random feedback filters \citep{lillicrap2016}.

\textbf{Dissociation between task performance and brain alignment.} Task accuracy and brain alignment are dissociated in our small-scale setting. At V1, the untrained baseline ($\NV{acc.rnd}\,\%$ test accuracy) exceeds BP ($\NV{acc.bp}\,\%$), and at IT, conditions with test accuracies between $\NV{acc.rnd}\,\%$ and $\NV{acc.bp}\,\%$ cannot be distinguished. This suggests that, at least for small CNNs trained on CIFAR-10, the relevant axis is not ``how well does it classify'' but ``what representational structure does the objective induce at each layer.'' Whether this dissociation persists at ImageNet scale, where task-optimized models achieve substantially higher Brain-Score alignment \citep{schrimpf2020}, remains an open question.

\textbf{Implications.} Our results carry two concrete methodological messages. For computational neuroscience: studies comparing trained models to brain data must always include an untrained baseline with the same architecture. Without this control, architecture effects are confounded with learning effects---our data show this confound can be essentially complete at V1. The same control must be run at the training resolution: the untrained network's lead over BP at V1 is present at $224$\,px evaluation but absent at the $32$\,px training resolution; it arises from the mismatch between training and evaluation resolution, not from training per se degrading early-layer structure (\newid). A second message comes from the revision itself: when subjects share a presentation order, same-run stimulus pairs can inflate between-subject agreement and the noise ceiling, and should be excluded or modelled.

\subsection{Limitations}

Several limitations bound our conclusions. (1)~All models were trained on CIFAR-10 using a small custom CNN, far below the scale at which ResNet-50 or ViT models achieve competitive Brain-Score results. The small architecture limits representational capacity and may contribute to the absence of differences at IT. (2)~We have only 3 subjects, with sub-03 showing consistently weak signal; our core findings are driven by sub-01 and sub-02. With three subjects only the lower noise bound is informative. Replication on a larger dataset (e.g., Natural Scenes Dataset, $N = 8$) would strengthen generalizability. (3)~STDP was applied to static images via Poisson spike trains, discarding the temporal dynamics that STDP was designed to exploit. (4)~The best-layer-per-ROI analysis reveals that FC1 sometimes produces higher V1 alignment than Conv1, complicating the anatomical mapping assumption. (5)~Gabor-peakedness scores show no significant differences across learning rules, but this metric captures only a coarse proxy for V1-like filter structure. (6)~Restricting the analysis to cross-run pairs removes $\NV{pairs.excl.pct}\,\%$ of the stimulus pairs; it removes the shared run component, but the sensitivity analyses (appendix) show that the main conclusions do not depend on this choice. (7)~We use a single dataset (THINGS-fMRI); cross-dataset replication would strengthen all claims. (8)~Models were trained at $32\times32$ resolution but evaluated at $224\times224$; the size of the untrained baseline's V1 advantage depends strongly on the evaluation resolution, which is quantified in \newid.

\section{Conclusion}

We presented a systematic RSA comparison of four learning rules plus an untrained baseline at $224\times224$ resolution, averaged across 5 seeds, against the THINGS-fMRI benchmark. In our small-scale setup, the learning rule has little effect on cortical alignment: architecture dominates early visual representations (V1/V2), where the untrained baseline exceeds all trained rules and, at V1, is the only condition that reaches the lower noise bound. This advantage holds at the $224$\,px evaluation resolution; in the model set of the resolution study (same architecture and training, 5 seeds) the V1 gap is $\Delta\rho = \NV{res.gap.224}$ (95\% CI $[\NV{res.gap.224.lo},\NV{res.gap.224.hi}]$) at $224$\,px, and when the models are evaluated at their $32$\,px training resolution it vanishes ($\Delta\rho = \NV{res.gap.32}$, 95\% CI $[\NV{res.gap.32.lo},\NV{res.gap.32.hi}]$): BP and the untrained network do not differ at V1, and the untrained network's lead at $224$\,px arises from the difference between training and evaluation resolution (\newid). At LOC, three of the four trained rules exceed the untrained baseline (FA borderline) without differing from each other; at IT, no two conditions can be distinguished at the low reliability of the data---though this may also reflect model capacity limitations. FA consistently produces the lowest alignment at V1 despite above-chance accuracy ($\NV{acc.fa}\,\%$). These results suggest that, at least at this scale, learning-rule comparisons at early areas are dominated by the architecture and by the relation between training and evaluation resolution, and should report both. Whether these findings generalize to larger architectures trained on richer datasets remains an important open question.

\section*{Acknowledgements}

The author thanks Blake Richards for the arXiv endorsement, the creators of the THINGS-fMRI dataset for making their data publicly available, and the Brain-Score team for their evaluation infrastructure. Analysis code and text drafts were developed with the assistance of an AI coding assistant; all results were verified by the author.

\section*{Code Availability}

Code and results: \url{https://github.com/nilsleut/learning-rules-rsa/tree/arxiv-v4}.

\appendix

\section{Sensitivity analyses}

\textbf{Pair set and block cleaning.} Table~\ref{tab:sens} gives the noise bounds and the per-subject model scores for three pair sets: all stimulus pairs (as in earlier versions), cross-run pairs (main analysis), and cross-run pairs with block cleaning, in which, separately for each subject, the mean of every run-pair block is subtracted from that subject's RDM and from the model RDM it is compared with. Block cleaning is defined only for the per-subject convention: the three subjects have different run structures, so no single cleaning applies to their average. The last column gives the original (pre-repair) model set on cross-run pairs. Restricting to cross-run pairs lowers the lower bound by $\NV{lbchg.V1}\,\%$ at V1, $\NV{lbchg.V2}\,\%$ at V2, $\NV{lbchg.LOC}\,\%$ at LOC and $\NV{lbchg.IT}\,\%$ at IT; block cleaning changes it little. The model scores change by at most $\NV{sens.maxdiff}$ between variants.

\begin{table*}[t]
\centering
\caption{Sensitivity: noise bounds and $\rho_{\text{subj}}$ (mean of 5 seeds) for all pairs, cross-run pairs (main), and cross-run pairs with per-subject block cleaning; repaired model set. Last column: original model set, cross-run pairs. Lower bound with stimulus-bootstrap 95\% CI. The upper bound is listed for completeness; it is not informative at $N=3$ (Methods).}
\label{tab:sens}
\small
\begin{tabular}{llcccc}
\toprule
ROI & Quantity & all pairs & cross-run & block-cleaned & cross-run, original set \\
\midrule
V1 & lower bound & $0.057\;[0.045,0.069]$ & $0.051\;[0.039,0.064]$ & $0.054\;[0.041,0.066]$ & -- \\
 & upper bound & $0.575$ & $0.572$ & $0.572$ & -- \\
 & Random & $0.053$ & $0.053$ & $0.053$ & $0.053$ \\
 & BP & $0.021$ & $0.022$ & $0.021$ & $0.024$ \\
 & FA & $0.007$ & $0.009$ & $0.006$ & $0.009$ \\
 & PC & $0.011$ & $0.012$ & $0.010$ & $0.039$ \\
 & STDP & $0.025$ & $0.027$ & $0.024$ & $0.045$ \\
\addlinespace
V2 & lower bound & $0.033\;[0.025,0.041]$ & $0.024\;[0.016,0.032]$ & $0.026\;[0.018,0.034]$ & -- \\
 & upper bound & $0.568$ & $0.563$ & $0.564$ & -- \\
 & Random & $0.029$ & $0.029$ & $0.028$ & $0.029$ \\
 & BP & $0.011$ & $0.012$ & $0.010$ & $0.013$ \\
 & FA & $0.003$ & $0.004$ & $0.002$ & $0.004$ \\
 & PC & $0.005$ & $0.006$ & $0.004$ & $0.019$ \\
 & STDP & $0.014$ & $0.015$ & $0.012$ & $0.024$ \\
\addlinespace
LOC & lower bound & $0.034\;[0.027,0.042]$ & $0.018\;[0.011,0.027]$ & $0.019\;[0.011,0.027]$ & -- \\
 & upper bound & $0.566$ & $0.558$ & $0.559$ & -- \\
 & Random & $-0.002$ & $-0.003$ & $-0.005$ & $-0.003$ \\
 & BP & $0.007$ & $0.008$ & $0.008$ & $0.008$ \\
 & FA & $0.003$ & $0.004$ & $0.003$ & $0.004$ \\
 & PC & $0.004$ & $0.004$ & $0.003$ & $0.003$ \\
 & STDP & $0.004$ & $0.004$ & $0.003$ & $0.004$ \\
\addlinespace
IT & lower bound & $0.035\;[0.028,0.042]$ & $0.021\;[0.014,0.028]$ & $0.021\;[0.014,0.028]$ & -- \\
 & upper bound & $0.563$ & $0.556$ & $0.556$ & -- \\
 & Random & $0.005$ & $0.004$ & $0.003$ & $0.004$ \\
 & BP & $0.008$ & $0.009$ & $0.009$ & $0.010$ \\
 & FA & $0.007$ & $0.008$ & $0.006$ & $0.008$ \\
 & PC & $0.007$ & $0.008$ & $0.007$ & $0.008$ \\
 & STDP & $0.008$ & $0.009$ & $0.008$ & $0.007$ \\
\bottomrule
\end{tabular}

\end{table*}

\textbf{Conclusions under each variant.} The statements of the main text (untrained exceeds BP at V1 and V2; BP exceeds the untrained network at LOC; FA lowest at V1; no significant differences among trained rules at LOC; none at IT) have the same sign and the same significance in all three pair sets, with one exception. In the repaired model set with block cleaning, BP vs.\ FA at LOC becomes borderline significant: $\NV{border.d}$ $[\NV{border.lo},\NV{border.hi}]$, against $\NV{border.cr.d}$ $[\NV{border.cr.lo},\NV{border.cr.hi}]$ on cross-run pairs without cleaning. We treat it as borderline.

\textbf{Layer assignment of V2} (Table~\ref{tab:v2conv2}). The fixed mapping assigns V2 to Conv1. With V2 assigned to Conv2 instead (same pair set, resamples and per-subject convention), the untrained network still exceeds BP, and it and STDP still exceed FA; BP falls below FA, and PC is no longer distinguishable from FA. The ordering among FA, BP and PC at V2 therefore depends on the layer assignment, whereas Random $>$ BP holds under both.

\begin{table}[h]
\centering
\caption{Differences at V2 ($\rho_{\text{subj}}(A) - \rho_{\text{subj}}(B)$, stimulus-bootstrap 95\% CI; cross-run pairs, repaired model set) with V2 assigned to Conv1 (main analysis) or to Conv2. Random $-$ BP is a primary comparison; the comparisons with FA at V2 are exploratory, because their ordering depends on the layer assignment.}
\label{tab:v2conv2}
\small
\resizebox{\columnwidth}{!}{%
\begin{tabular}{lccc}
\toprule
Difference at V2 & Conv1$\to$V2 (main) & Conv2$\to$V2 & same sign and significance \\
\midrule
Random $-$ BP & $+0.0165\;[0.0072,0.0261]$ & $+0.0526\;[0.0401,0.0656]$ & yes \\
Random $-$ FA & $+0.0252\;[0.0150,0.0361]$ & $+0.0313\;[0.0212,0.0413]$ & yes \\
BP $-$ FA & $+0.0087\;[0.0053,0.0125]$ & $-0.0213\;[-0.0274,-0.0148]$ & no \\
PC $-$ FA & $+0.0021\;[0.0005,0.0037]$ & $-0.0005\;[-0.0035,0.0027]$ & no \\
STDP $-$ FA & $+0.0110\;[0.0077,0.0146]$ & $+0.0097\;[0.0047,0.0152]$ & yes \\
\bottomrule
\end{tabular}
}
\end{table}

\textbf{Correction for multiple comparisons} (Table~\ref{tab:fdr}). Two-sided bootstrap $p$-values from the same $\NV{nboot}$ resamples, $p = (2\min(\#\{d_b \le 0\}, \#\{d_b \ge 0\}) + 1)/(B + 1)$, corrected by Benjamini--Hochberg (false discovery rate) and by Holm (family-wise error rate), each at the 5\,\% level, within the current family of $\NV{nprimary}$ primary comparisons and within the earlier family of $\NV{nprimary.old}$, which also contained FA against every other condition at V2. All $\NV{nprimary.old}$ comparisons of the earlier family, including the four at V2, are significant under both corrections, and all $\NV{nprimary}$ of the current family under both corrections in both families.

\begin{table*}[t]
\centering
\caption{All $\NV{nprimary.old}$ comparisons of the earlier primary family with bootstrap $p$-values and adjusted $p$-values (Benjamini--Hochberg, Holm) within the current family of $\NV{nprimary}$ (above the line) and within the earlier family of $\NV{nprimary.old}$ (all rows; the four below the line belonged to the earlier family only). Per-subject convention, cross-run pairs, repaired model set. The smallest attainable bootstrap $p$ is $1/(B+1)$.}
\label{tab:fdr}
\small
\begin{tabular}{llcccccc}
\toprule
Comparison & ROI & $\Delta\rho_{\text{subj}}$ [95\% CI] & $p_{\text{boot}}$ & BH, 7 & BH, 11 & Holm, 7 & Holm, 11 \\
\midrule
Random $-$ BP & V1 & $+0.0308\;[0.0186,0.0428]$ & $0.0010$ & $0.0014$ & $0.0014$ & $0.0070$ & $0.0110$ \\
Random $-$ BP & V2 & $+0.0165\;[0.0072,0.0261]$ & $0.0010$ & $0.0014$ & $0.0014$ & $0.0070$ & $0.0110$ \\
Random $-$ BP & LOC & $-0.0110\;[-0.0206,-0.0023]$ & $0.0190$ & $0.0190$ & $0.0190$ & $0.0190$ & $0.0190$ \\
Random $-$ FA & V1 & $+0.0446\;[0.0313,0.0580]$ & $0.0010$ & $0.0014$ & $0.0014$ & $0.0070$ & $0.0110$ \\
BP $-$ FA & V1 & $+0.0138\;[0.0092,0.0186]$ & $0.0010$ & $0.0014$ & $0.0014$ & $0.0070$ & $0.0110$ \\
FA $-$ PC & V1 & $-0.0033\;[-0.0054,-0.0011]$ & $0.0030$ & $0.0035$ & $0.0037$ & $0.0070$ & $0.0110$ \\
FA $-$ STDP & V1 & $-0.0179\;[-0.0229,-0.0130]$ & $0.0010$ & $0.0014$ & $0.0014$ & $0.0070$ & $0.0110$ \\
\midrule
Random $-$ FA & V2 & $+0.0252\;[0.0150,0.0361]$ & $0.0010$ & -- & $0.0014$ & -- & $0.0110$ \\
BP $-$ FA & V2 & $+0.0087\;[0.0053,0.0125]$ & $0.0010$ & -- & $0.0014$ & -- & $0.0110$ \\
FA $-$ PC & V2 & $-0.0021\;[-0.0037,-0.0005]$ & $0.0090$ & -- & $0.0099$ & -- & $0.0180$ \\
FA $-$ STDP & V2 & $-0.0110\;[-0.0146,-0.0077]$ & $0.0010$ & -- & $0.0014$ & -- & $0.0110$ \\
\bottomrule
\end{tabular}

\end{table*}

\textbf{Diagnostics of the run structure} (Table~\ref{tab:diag}). If the run component were shared beyond same-run pairs, the mean dissimilarity of each pair of runs would agree between sub-01 and sub-02. The Spearman correlation of their 45 run-pair means is small and not significant at any ROI (exact one-sided $p$ over all $\NV{rp.nperm}$ relabellings of sub-02's runs). With only 45 run pairs this test has little power, so block cleaning is reported as a robustness check rather than ruled out. Block cleaning changes the V1 luminance fit by $\NV{lum.s3.V1.chg}\,\%$ for sub-03, whose run order differs, and by $\NV{lum.s1.V1.chg}\,\%$ (sub-01) and $\NV{lum.s2.V1.chg}\,\%$ (sub-02); the pre-specified stop criterion (a decrease of more than 30\,\% for sub-01 or sub-02) is not met.

\begin{table}[h]
\centering
\caption{Diagnostics of the run structure. Top: agreement of the run-pair mean dissimilarities between sub-01 and sub-02 (identical presentation order). Bottom: V1 luminance fit per subject on cross-run pairs before and after block cleaning.}
\label{tab:diag}
\small
\resizebox{\columnwidth}{!}{%
\begin{tabular}{lcc}
\toprule
ROI & $\rho$ (sub-01 vs sub-02, 45 run-pair means) & $p$ (exact, one-sided) \\
\midrule
V1 & $0.055$ & $0.336$ \\
V2 & $0.065$ & $0.323$ \\
LOC & $0.171$ & $0.160$ \\
IT & $0.097$ & $0.226$ \\
\midrule
Subject & \multicolumn{2}{c}{V1 luminance fit: cross-run $\to$ block-cleaned (change)} \\
\midrule
sub-01 & \multicolumn{2}{c}{$0.054 \to 0.054$ ($-1\,\%$)} \\
sub-02 & \multicolumn{2}{c}{$0.096 \to 0.086$ ($-10\,\%$)} \\
sub-03 & \multicolumn{2}{c}{$0.011 \to 0.011$ ($-2\,\%$)} \\
\bottomrule
\end{tabular}
}
\end{table}

\begin{table*}[t]
\centering
\caption{RSA scores in the mean-RDM convention $\rho_{\text{mean}}$ (model RDM vs.\ the average of the three subject RDMs; the convention of earlier versions), with stimulus-bootstrap 95\% CI; cross-run pairs, repaired model set, mean of 5 seeds. Bold, $^{\dagger}$, $^{\ddagger}$ as in Table~\ref{tab:rsa}. The lower bound is defined per subject and is given in Table~\ref{tab:rsa}.}
\label{tab:rsa_mr}
\small
\resizebox{\textwidth}{!}{\begin{tabular}{llcccccc}
\toprule
ROI & Layer & lower bound & Random & BP & FA & PC & STDP \\
\midrule
V1 & Conv1 & -- & $\mathbf{0.077}\;[.055,.099]$ & $0.034\;[.020,.050]^{\dagger}$ & $0.014\;[.000,.028]^{\dagger}$ & $0.018\;[.006,.034]^{\dagger}$ & $0.039\;[.025,.056]^{\dagger}$ \\
V2 & Conv1 & -- & $\mathbf{0.043}\;[.027,.061]$ & $0.019\;[.007,.031]^{\dagger}$ & $0.005\;[-.006,.017]^{\dagger}$ & $0.009\;[-.003,.020]^{\dagger}$ & $0.022\;[.010,.035]^{\dagger}$ \\
LOC & Conv3 & -- & $-0.007\;[-.018,.004]$ & $\mathbf{0.013}\;[.001,.027]^{\dagger}$ & $0.006\;[-.006,.019]^{\dagger}$ & $0.007\;[-.004,.018]^{\dagger}$ & $0.007\;[-.004,.017]^{\dagger}$ \\
IT & FC1 & -- & $0.006\;[-.003,.016]$ & $0.015\;[.006,.024]$ & $0.014\;[.004,.024]$ & $0.013\;[.003,.023]$ & $0.015\;[.005,.024]$$^{\ddagger}$ \\
\bottomrule
\end{tabular}
}
\end{table*}

\begin{table*}[t]
\centering
\caption{All pairwise differences $\rho_{\text{subj}}(A) - \rho_{\text{subj}}(B)$ with stimulus-bootstrap 95\% CI (cross-run pairs, repaired model set). Bold: CI excludes 0. $^{\mathrm{P}}$: primary comparison (Methods); all others are exploratory, without correction for multiple testing.}
\label{tab:pairwise}
\small
\resizebox{\textwidth}{!}{\begin{tabular}{lcccc}
\toprule
Pair (A $-$ B) & V1 & V2 & LOC & IT \\
\midrule
Random $-$ BP & $\mathbf{+0.0308}\;[0.0186,0.0428]^{\mathrm{P}}$ & $\mathbf{+0.0165}\;[0.0072,0.0261]^{\mathrm{P}}$ & $\mathbf{-0.0110}\;[-0.0206,-0.0023]^{\mathrm{P}}$ & $-0.0054\;[-0.0144,0.0036]$ \\
Random $-$ FA & $\mathbf{+0.0446}\;[0.0313,0.0580]^{\mathrm{P}}$ & $\mathbf{+0.0252}\;[0.0150,0.0361]$ & $-0.0066\;[-0.0140,0.0002]$ & $-0.0039\;[-0.0118,0.0039]$ \\
Random $-$ PC & $\mathbf{+0.0413}\;[0.0292,0.0533]$ & $\mathbf{+0.0231}\;[0.0140,0.0327]$ & $\mathbf{-0.0073}\;[-0.0137,-0.0013]$ & $-0.0037\;[-0.0115,0.0041]$ \\
Random $-$ STDP & $\mathbf{+0.0267}\;[0.0154,0.0381]$ & $\mathbf{+0.0142}\;[0.0051,0.0238]$ & $\mathbf{-0.0073}\;[-0.0138,-0.0010]$ & $-0.0048\;[-0.0123,0.0028]$ \\
BP $-$ FA & $\mathbf{+0.0138}\;[0.0092,0.0186]^{\mathrm{P}}$ & $\mathbf{+0.0087}\;[0.0053,0.0125]$ & $+0.0044\;[-0.0011,0.0102]$ & $+0.0016\;[-0.0041,0.0075]$ \\
BP $-$ PC & $\mathbf{+0.0104}\;[0.0065,0.0149]$ & $\mathbf{+0.0066}\;[0.0035,0.0101]$ & $+0.0037\;[-0.0017,0.0097]$ & $+0.0017\;[-0.0036,0.0072]$ \\
BP $-$ STDP & $\mathbf{-0.0042}\;[-0.0079,-0.0004]$ & $-0.0023\;[-0.0050,0.0007]$ & $+0.0037\;[-0.0017,0.0097]$ & $+0.0006\;[-0.0043,0.0056]$ \\
FA $-$ PC & $\mathbf{-0.0033}\;[-0.0054,-0.0011]^{\mathrm{P}}$ & $\mathbf{-0.0021}\;[-0.0037,-0.0005]$ & $-0.0007\;[-0.0032,0.0018]$ & $+0.0001\;[-0.0041,0.0044]$ \\
FA $-$ STDP & $\mathbf{-0.0179}\;[-0.0229,-0.0130]^{\mathrm{P}}$ & $\mathbf{-0.0110}\;[-0.0146,-0.0077]$ & $-0.0007\;[-0.0042,0.0030]$ & $-0.0009\;[-0.0050,0.0029]$ \\
PC $-$ STDP & $\mathbf{-0.0146}\;[-0.0186,-0.0105]$ & $\mathbf{-0.0089}\;[-0.0119,-0.0062]$ & $+0.0000\;[-0.0030,0.0030]$ & $-0.0011\;[-0.0045,0.0022]$ \\
\bottomrule
\end{tabular}
}
\end{table*}

\section{Errata of earlier versions}

\textbf{Accuracy.} Table~\ref{tab:accold} gives the accuracies printed in earlier versions next to the test accuracies of this version. The earlier values are accuracies on training batches in the last epoch of the original runs, measured with batch normalisation in training mode; for PC and STDP this is also the mode in which the evaluation defect left them. They are not test accuracies and are not comparable to Table~\ref{tab:accuracy}.

\begin{table}[h]
\centering
\caption{Accuracy printed in v2 (training batches, original runs) and CIFAR-10 test accuracy of the repaired models (seed~0).}
\label{tab:accold}
\small
\resizebox{\columnwidth}{!}{\begin{tabular}{lcc}
\toprule
Condition & v2: training accuracy (\%) & v4: test accuracy (\%) \\
\midrule
Backprop & $82.4$ & $75.0$ \\
STDP & $63.2$ & $57.4$ \\
Predictive Coding & $56.6$ & $56.7$ \\
Feedback Alignment & $39.0$ & $38.8$ \\
Random Weights & $10.0$ & $10.0$ \\
\bottomrule
\end{tabular}
}
\end{table}

Table~\ref{tab:errata} lists values printed in v2 next to the value its own source gives. The first four are double-rounding errors (rounding first to four, then to three decimals). The V1 gap printed in v2 ($+0.044$) is that of the resolution study (\newid, $\NV{err.gap.sweep}$), while the table it accompanied gives $\NV{err.gap.V1}$. The noise bounds printed in v2 (upper 0.11, 0.09, 0.06, 0.07; lower 0.07, 0.05, 0.03, 0.04) could not be traced to any stored result and are replaced by the bounds above.

\begin{table*}[t]
\centering
\caption{Errata of v2: printed value and the value of its source (original analysis).}
\label{tab:errata}
\small
\begin{tabular}{llll}
\toprule
Location in v2 & Quantity & printed & correct \\
\midrule
Abstract, \S4.2, Tab.~2, \S5 & Random, V1 & 0.076 & $0.075$ \\
Abstract, \S4.2, Tab.~2, \S5 & BP, V1 & 0.034 & $0.033$ \\
Tab.~2 & BP, V2 & 0.019 & $0.018$ \\
Tab.~2 & BP, V2, CI upper & .023 & $.022$ \\
Abstract, \S1, \S4.2, \S5 & $\Delta\rho$ Random $-$ BP, V1 & +0.044 & $+0.042$ \\
\S4.8 & Gabor peakedness, BP & 2.00 & $1.99$ \\
\S4.5 & seed SD range, V1/V2 & 0.003--0.007 & $0.003$--$0.011$ \\
\S4.4 & FDR-significant comparisons of 40 & 30 & $26$ \\
\S4.4 & largest FDR $p$ at V1/V2 & $p \le 0.014$ & $0.0144$ \\
\bottomrule
\end{tabular}

\end{table*}

\bibliographystyle{apalike}

\begin{thebibliography}{99}

\bibitem[Bi and Poo, 1998]{bi1998}
Bi, G.-q. and Poo, M.-m. (1998).
Synaptic modifications in cultured hippocampal neurons: dependence on spike timing, synaptic strength, and postsynaptic cell type.
\textit{Journal of Neuroscience}, 18:10464--10472.

\bibitem[Crick, 1989]{crick1989}
Crick, F. (1989).
The recent excitement about neural networks.
\textit{Nature}, 337:129--132.

\bibitem[Hebart et~al., 2019]{hebart2019}
Hebart, M.~N., Dickter, A.~H., Kidder, A., Kwok, W.~Y., Corriveau, A., Van~Wicklin, C., and Baker, C.~I. (2019).
THINGS: A database of 1,854 object concepts and more than 26,000 naturalistic object images.
\textit{PLOS ONE}, 14:e0223792.

\bibitem[Hebart et~al., 2023]{hebart2023}
Hebart, M.~N., Contier, O., Teichmann, L., Rockter, A.~H., Zheng, C.~Y., Kidder, A., Corriveau, A., Vaziri-Pashkam, M., and Baker, C.~I. (2023).
THINGS-data, a multimodal collection of large-scale datasets for investigating object representations in human brain and behavior.
\textit{eLife}, 12:e82580. doi:10.7554/eLife.82580.

\bibitem[Kriegeskorte et~al., 2008]{kriegeskorte2008}
Kriegeskorte, N., Mur, M., and Bandettini, P. (2008).
Representational similarity analysis -- connecting the branches of systems neuroscience.
\textit{Frontiers in Systems Neuroscience}, 2:4.

\bibitem[Lillicrap et~al., 2016]{lillicrap2016}
Lillicrap, T.~P., Cownden, D., Tweed, D.~B., and Akerman, C.~J. (2016).
Random synaptic feedback weights support error backpropagation for deep learning.
\textit{Nature Communications}, 7:13276.

\bibitem[Lillicrap et~al., 2020]{lillicrap2020}
Lillicrap, T.~P., Santoro, A., Marris, L., Akerman, C.~J., and Hinton, G. (2020).
Backpropagation and the brain.
\textit{Nature Reviews Neuroscience}, 21:335--346.

\bibitem[Nili et~al., 2014]{nili2014}
Nili, H., Wingfield, C., Walther, A., Su, L., Marslen-Wilson, W., and Kriegeskorte, N. (2014).
A toolbox for representational similarity analysis.
\textit{PLoS Computational Biology}, 10:e1003553.

\bibitem[Olshausen and Field, 1996]{olshausen1996}
Olshausen, B.~A. and Field, D.~J. (1996).
Emergence of simple-cell receptive field properties by learning a sparse code for natural images.
\textit{Nature}, 381:607--609.

\bibitem[Rao and Ballard, 1999]{rao1999}
Rao, R.~P.~N. and Ballard, D.~H. (1999).
Predictive coding in the visual cortex: a functional interpretation of some extra-classical receptive-field effects.
\textit{Nature Neuroscience}, 2:79--87.

\bibitem[Saxe et~al., 2011]{saxe2011}
Saxe, A.~M., Koh, P.~W., Chen, Z., Bhand, M., Suresh, B., and Ng, A.~Y. (2011).
On random weights and unsupervised feature learning.
In \textit{Proceedings of the 28th International Conference on Machine Learning}.

\bibitem[Schrimpf et~al., 2020]{schrimpf2020}
Schrimpf, M., Kubilius, J., Hong, H., Majaj, N.~J., Rajalingham, R., Issa, E.~B., Kar, K., Bashivan, P., Prescott-Roy, J., Schmidt, K., Yamins, D.~L.~K., and DiCarlo, J.~J. (2020).
Brain-Score: Which artificial neural network for object recognition is most brain-like?
\textit{bioRxiv}.

\bibitem[Truzzi and Cusack, 2025]{truzzi2025}
Truzzi, A. and Cusack, R. (2025).
Neural responses in early visual cortex are well predicted by random-weight CNNs.
\textit{bioRxiv}.

\bibitem[Walther et~al., 2016]{walther2016}
Walther, A., Nili, H., Ejaz, N., Alink, A., Kriegeskorte, N., and Diedrichsen, J. (2016).
Reliability of dissimilarity measures for multi-voxel pattern analysis.
\textit{NeuroImage}, 137:188--200. doi:10.1016/j.neuroimage.2015.12.012.

\bibitem[Whittington and Bogacz, 2017]{whittington2017}
Whittington, J.~C.~R. and Bogacz, R. (2017).
An approximation of the error backpropagation algorithm in a predictive coding network with local Hebbian synaptic plasticity.
\textit{Neural Computation}, 29:1229--1262.

\bibitem[Yamins and DiCarlo, 2016]{yamins2016}
Yamins, D.~L.~K. and DiCarlo, J.~J. (2016).
Using goal-driven deep learning models to understand sensory cortex.
\textit{Nature Neuroscience}, 19:356--365.

\end{thebibliography}

\end{document}